\documentclass[conference]{IEEEtran}
\IEEEoverridecommandlockouts
\usepackage{cite}
\usepackage{amsmath,amssymb,amsfonts}
\usepackage{algorithmic}
\usepackage{graphicx}
\usepackage{textcomp}
\usepackage{colortbl,booktabs}
\usepackage{float}
\usepackage{xcolor}
\def\BibTeX{{\rm B\kern-.05em{\sc i\kern-.025em b}\kern-.08em
    T\kern-.1667em\lower.7ex\hbox{E}\kern-.125emX}}
\begin{document}

\title{USIS-PGM: Photometric Gaussian Mixtures for Underwater Salient Instance Segmentation\\

\thanks{*The work described in this paper was partially supported by grants AoE/E-601/24-N, 16203223, C6029-23G,  and C6078-25G from the Research Grants Council of the Hong Kong Special Administrative Region, China.}
}

\author{\IEEEauthorblockN{1\textsuperscript{st} Lin Hong}
\IEEEauthorblockA{\textit{Department of Electronic and Computer Engineering} \\
\textit{The Hong Kong University of Science and Technology}\\
Hong Kong, China \\
eelinhong@ust.hk}
\and
\IEEEauthorblockN{2\textsuperscript{nd} Xiangtong Yao}
\IEEEauthorblockA{\textit{School of CIT, Chair i6} \\
\textit{Technical University of Munich}\\
Munich, Germany \\
xiangtong.yao@tum.de}
\and
\IEEEauthorblockN{3\textsuperscript{rd} Mürüvvet Bozkurt}
\IEEEauthorblockA{\textit{School of CIT} \\
\textit{Technical University of Munich}\\
Munich, Germany \\
mbozkurtt76@gmail.com}
\and

\IEEEauthorblockN{4\textsuperscript{th} Xin Wang}
\IEEEauthorblockA{\textit{School of Robotics and Advanced Manufacture} \\
\textit{Harbin Institute of Technology (Shenzhen)}\\
Shenzhen, China \\
wangxinsz@hit.edu.cn}
\and

\IEEEauthorblockN{5\textsuperscript{th} Fumin Zhang}
\IEEEauthorblockA{\textit{Department of Electronic and Computer Engineering} \\
\textit{The Hong Kong University of Science and Technology}\\
Hong Kong, China \\
eefumin@ust.hk}
}
\maketitle

\begin{abstract}
Underwater salient instance segmentation (USIS) is crucial for marine robotic systems, as it enables both underwater salient object detection and instance-level mask prediction for visual scene understanding. Compared with its terrestrial counterpart, USIS is more challenging due to the underwater image degradation. To address this issue, this paper proposes USIS-PGM, a single-stage framework for USIS. Specifically, the encoder enhances boundary cues through a frequency-aware module and performs content-adaptive feature reweighting via a dynamic weighting module. The decoder incorporates a Transformer-based instance activation module to better distinguish salient instances. In addition, USIS-PGM employs multi-scale Gaussian heatmaps generated from ground-truth masks through Photometric Gaussian Mixture (PGM) to supervise intermediate decoder features, thereby improving salient instance localization and producing more structurally coherent mask predictions. Experimental results demonstrate the superiority and practical applicability of the proposed USIS-PGM model.
\end{abstract}

\section{Introduction}
\begin{figure*}[t]
    \centering
    \includegraphics[width=0.99\textwidth]{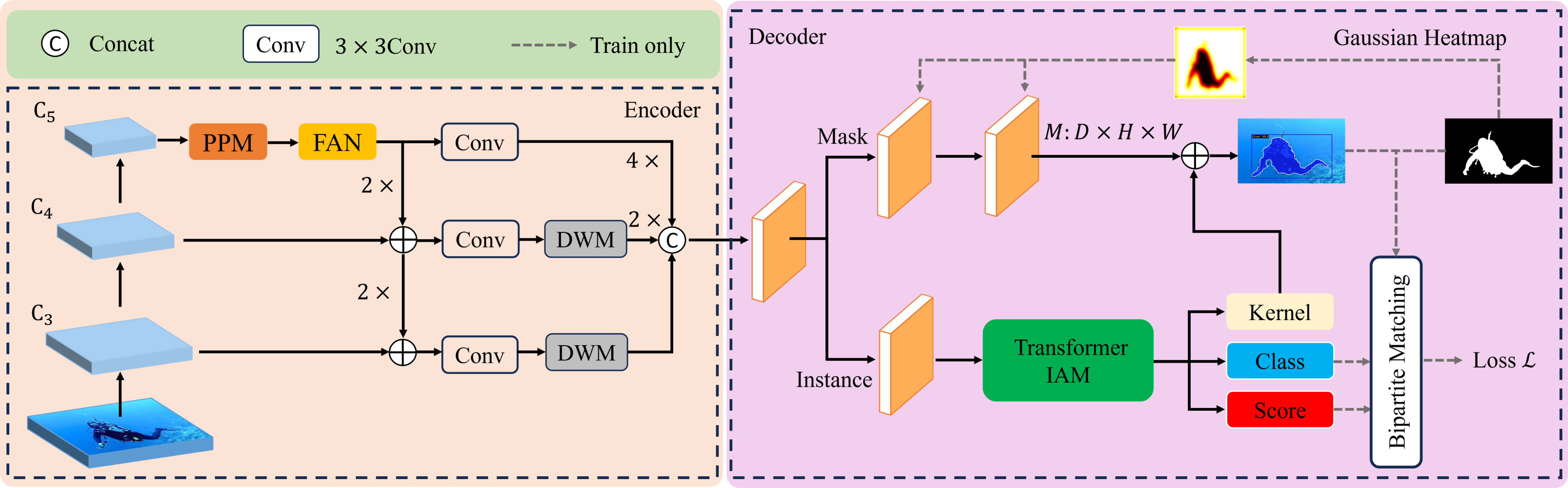}
\caption{Overall framework of the proposed USIS-PGM model, consisting of an encoder, a decoder, and the PGM-based multi-scale supervision strategy. The encoder integrates FAN and DWM to enhance boundary-aware representation and content-adaptive feature reweighting, while the decoder employs a Transformer-based IAM for accurate salient instance localization and mask prediction.}
    \label{fig:framework}
\end{figure*}
Visual perception plays a critical role in underwater robotic systems for a wide range of applications, including marine exploration~\cite{mallet2014underwater}, biodiversity monitoring~\cite{huang2023underwater}, and underwater infrastructure inspection~\cite{hong2023vision}. In these scenarios, underwater robots are often required not only to detect visually attractive objects, but also to distinguish individual object instances for downstream decision-making and interaction. Underwater Salient Instance Segmentation (USIS)~\cite{lian2024diving,hong2025usis16k}, which jointly addresses \emph{where to look} (saliency prediction) and \emph{what is there} (instance segmentation) in underwater scenes, has therefore emerged as a fundamental task for underwater visual scene understanding. By simultaneously performing underwater salient object detection~\cite{hong2023usod10k} and predicting their instance-level masks, USIS is particularly valuable for visual perception of underwater robotic systems in complex environments.

Despite its importance, USIS remains highly challenging. In contrast to terrestrial scenes, underwater imagery is severely affected by absorption and scattering of light, wavelength-dependent attenuation, and non-uniform illumination. These factors often lead to low contrast, color distortion, and blurred boundaries, which significantly degrade the discriminability of visual features~\cite{hong2021wsuie}. Moreover, the underwater environment is characterized by dynamic backgrounds, scale variations among targets. Such characteristics make it difficult to accurately localize underwater salient objects and predict their precise masks from images. 

In terrestrial environments, salient instance segmentation has been widely studied and has achieved notable progress~\cite{fan2019s4net,wu2021regularized,pei2024calibnet}. However, these methods cannot be directly transferred to underwater environments in an off-the-shelf manner. The domain gap between terrestrial and underwater imagery is substantial, not only because of the distinct imaging physics, but also due to differences in object categories, visual appearance, and scene composition. In particular, object categories in underwater imagery often follow long-tailed distributions, with some classes appearing frequently while many others are sparsely represented. This further limits the generalization of salient instance segmentation models originally developed for terrestrial environments.

To address these challenges, existing research on USIS has mainly progressed along two directions. The first direction focuses on constructing large-scale datasets to support model development and evaluation. Representative examples include USIS10K~\cite{lian2024diving} dataset and USIS16K~\cite{hong2025usis16k} dataset. Specifically, USIS10K contains 10,632 underwater images with pixel-level annotations across 7 categories collected from diverse underwater scenes, while USIS16K includes 16,151 underwater images with multi-level annotations and covers 158 object categories. These datasets have significantly advanced the study of USIS by providing stronger benchmark support.

The second direction aims to design more robust models for USIS that explicitly account for underwater image degradation and the cross-domain gap between terrestrial and underwater scenes. \cite{lian2024diving} provided an underwater-specific baseline, termed USIS-SAM, for USIS10K by adapting the Segment Anything Model~\cite{kirillov2023segment}. It introduced an underwater-aware vision transformer encoder to incorporate underwater-domain visual prompts into the segmentation network. Although such efforts demonstrate the promise of underwater-oriented architectural adaptation, existing methods still face notable limitations. In particular, under severe degradation, intermediate features in the network can be easily distracted by low-level photometric corruption and background noise, which often leads to inaccurate salient instance localization, fragmented masks, and poor separation between adjacent instances. This suggests that, beyond simply strengthening backbone feature extraction, it is also important to impose more effective structural guidance on intermediate representations during decoding.

In this work, we argue that \emph{Photometric Gaussian Mixture} (PGM)~\cite{crombez2018visual} offers a promising solution to this problem. The central intuition is that Gaussian-distributed photometric priors provide compact yet informative guidance for modeling the spatial distribution, extent, and structural organization of salient instances. Compared with direct binary supervision, PGM-based supervision yields smoother and more ambiguity-tolerant training signals, which are particularly suitable for underwater scenes where object boundaries are often blurred and appearance cues are unstable. By incorporating multi-scale Gaussian heatmap guidance derived from ground-truth masks, the network is encouraged to learn more discriminative and spatially coherent intermediate representations, thereby improving both salient instance localization and mask prediction.  

Motivated by this observation, this paper proposes \textbf{USIS-PGM}, a dedicated USIS model with single-stage architecture that explicitly incorporates \emph{Photometric Gaussian Mixture} (PGM)-based supervision. Specifically, the encoder enhances boundary-aware representations through a frequency-aware module and performs content-adaptive feature reweighting via a dynamic weighting module. On the decoder side, a Transformer-based instance activation module is introduced to better distinguish salient instances and improve instance separation. Furthermore, intermediate decoder features are explicitly supervised by multi-scale Gaussian heatmaps generated from ground-truth masks.
Extensive experiments on the USIS16K dataset demonstrate that the proposed USIS-PGM model outperforms existing methods. The results further show that incorporating PGM-based supervision not only improves salient object localization, but also enhances mask completeness and structural consistency under underwater conditions.
The contributions of this paper are summarized as follows:
\begin{itemize}
    \item We propose \textbf{USIS-PGM}, a USIS model with single-stage architecture. It explicitly introduces PGM to generate multi-scale Gaussian heatmaps to guide intermediate decoder features and promote more effective learning of salient instance localization and mask prediction.
    
    \item The proposed USIS-PGM integrates frequency-aware boundary enhancement, dynamic feature reweighting, and Transformer-based instance activation into a unified architecture to better address underwater image degradation and improve salient instance segmentation.
   
    \item Extensive evaluations on the USIS16K dataset demonstrate the effectiveness and superiority of the proposed USIS-PGM model over benchmark methods.
\end{itemize}

\section{USIS-PGM model}

The overall architecture of the proposed USIS-PGM model is illustrated in Fig.~\ref{fig:framework}. USIS-PGM adopts a single-stage framework and consists of three main components: an encoder, a decoder, and a PGM-guided multi-scale supervision.

\subsection{Encoder Design}
Underwater imagery often suffers from low contrast and blurred boundaries. These factors make it difficult for standard feature extractors to preserve both semantic consistency and local structural details. To address this issue, the encoder incorporates two modules, i.e., frequency adjust network (FAN), and dynamic weighting module (DWM).


\textbf{FAN} is designed to enhance fine edges and contour details by reweighting the frequency content of feature maps. Its main motivation is that low-frequency components primarily describe global intensity and coarse structural information, whereas high-frequency components contain boundary, edge, and texture cues that are especially important for accurate object delineation. This distinction is particularly relevant in underwater scenes, where image degradation often weakens boundary visibility and makes salient instances difficult to separate from the background.
Based on this observation, FAN first transforms the input feature map into the frequency domain and emphasizes the high-frequency responses that are more closely related to edges and local structural variations. The selected frequency components are then converted back into the spatial domain to produce a gain map, which highlights regions with strong boundary and texture information. Finally, this gain map is applied to the original feature map in a residual manner, so that informative edge-sensitive responses are enhanced while the original low-frequency content and overall semantic structure are preserved.

\textbf{DWM} is designed to improve feature adaptivity by learning input-dependent weights to modulate feature responses. Specifically, the input feature map is first compressed through two consecutive convolutional layers to obtain a compact representation with reduced channel dimensionality. The compressed feature is then flattened, and dynamic modulation weights are generated from the feature content itself, enabling content-aware reweighting of individual feature elements. After modulation, the feature is reshaped back into the spatial domain and projected to the original channel dimension.
In our implementation, DWM is applied to the higher-level encoder branches $C_4$ and $C_5$. To reduce computational cost, these branches are first resized to a fixed spatial resolution, processed by DWM, and then upsampled back to their original size. Through this content-adaptive reweighting mechanism, DWM helps the network emphasize salient-instance-related information while suppressing noisy and irrelevant responses.

\subsection{Decoder with Instance Activation Module}
The decoder is designed to progressively reconstruct dense masks. One of the challenges in USIS is to accurately separate nearby or partially overlapping salient objects, particularly when their appearance is degraded and their boundaries are ambiguous. To address this challenge, we incorporate a Transformer-based Instance Activation Module (IAM) into the decoder.
The key idea of IAM is to enhance instance-aware feature interaction and activation~\cite{cheng2022sparse}. Unlike convolution-based decoding, the Transformer~\cite{vaswani2017attention} is capable of capturing long-range dependencies and modeling global relationships across spatial regions. This property is especially beneficial in underwater scenes, where local visual evidence alone is often insufficient to distinguish adjacent instances under blur, occlusion, or low contrast. By exploiting self-attention, IAM dynamically aggregates contextual cues from different locations and highlights features that are informative for salient instance separation.
Specifically, IAM takes decoder features as input and refines them through instance-aware attention modeling. This process promotes better disentanglement of features associated with different salient regions while suppressing ambiguous background responses.

\subsection{PGM-based Multi-scale Supervision}
To improve intermediate decoder representation learning, we introduce a multi-scale supervision strategy based on PGM. The underlying motivation is that binary mask supervision imposed only at the final output layer may be insufficient to guide intermediate features toward accurate salient-instance localization and structural modeling, especially in underwater images with blurred boundaries and unstable visual appearance. In contrast, Gaussian heatmaps generated from ground-truth masks provide smoother and more spatially structured supervisory signals, making them better suited to progressively guiding feature learning across different decoding stages.

A two-dimensional Gaussian function provides a natural way to model spatial influence that gradually decays with distance from a center location. Let $x \in \mathbb{R}$ and $y \in \mathbb{R}$ denote two spatial variables, and let $\mathbf{x}=(x,y)^\top$. The two-dimensional Gaussian function is defined as
\begin{equation}
f(x,y)=A\exp\left(
-\frac{(x-x_0)^2}{2\sigma_x^2}
-\frac{(y-y_0)^2}{2\sigma_y^2}
\right),
\label{eq:2d_gaussian}
\end{equation}
where $A$ denotes the amplitude, $(x_0,y_0)$ represents the Gaussian center, and $\sigma_x$ and $\sigma_y$ control the spread along the horizontal and vertical directions, respectively.

\begin{figure}[t]
    \centering
    \includegraphics[width=0.45\textwidth]{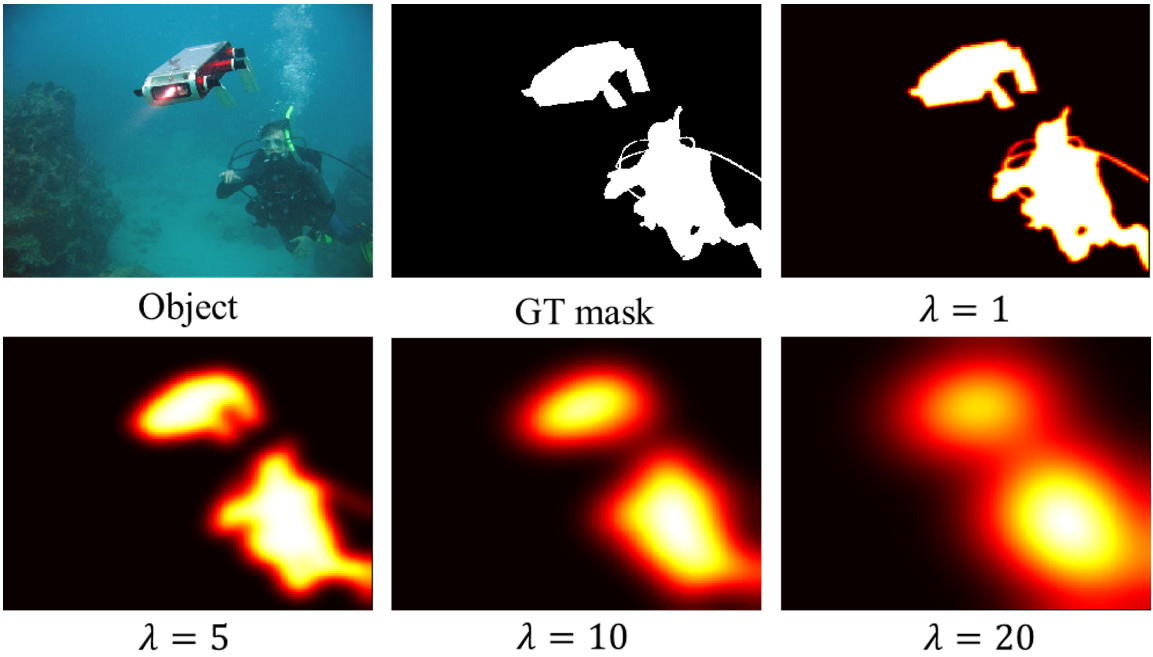}
    \caption{Visualization of Gaussian heatmaps generated from ground-truth mask based on $G(I;\lambda)$, where $\lambda=\{1,5,10,20\}$.}
    \label{fig:GPM}
\end{figure}
\begin{figure*}[t]
\begin{center}
\includegraphics[width=1\linewidth]{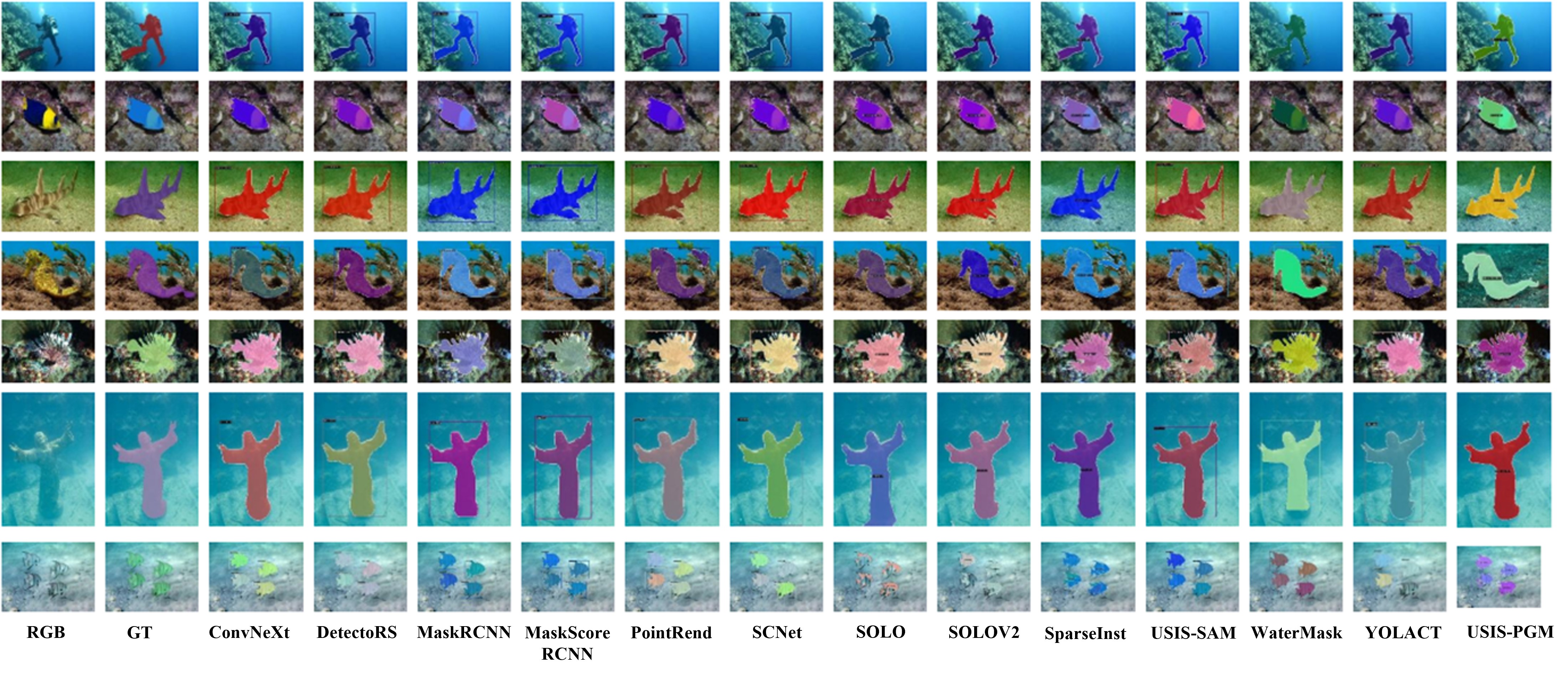}
\end{center}
    \caption{Qualitative evaluation of the proposed USIS-PGM model with 12 benchmark methods on the USIS16K dataset.}
\label{fig:qualitative_com}
\end{figure*} 
\textbf{Photometric Gaussian Mixture (PGM).}
Let an image be defined on the pixel grid $\mathcal{U}$, where each pixel coordinate is denoted by $\mathbf{u}=(u,v)^\top$ and the corresponding luminance value is $I(\mathbf{u})$. Following~\cite{crombez2018visual}, PGM is formulated as
\begin{equation}
G(I;\lambda)(\mathbf{x})
=\sum_{\mathbf{u}\in\mathcal{U}} I(\mathbf{u})\,
\exp\left(-\frac{\|\mathbf{x}-\mathbf{u}\|^2}{2\lambda^2}\right),
\label{eq:photometric_gmm}
\end{equation}
where $\lambda$ is a shared isotropic extension parameter that controls the spatial spread of each Gaussian component.

In our setting, the Gaussian center is aligned with the pixel position, i.e., $(x_0,y_0)=\mathbf{u}$, the amplitude is defined by the pixel intensity, i.e., $A=I(\mathbf{u})$, and isotropic extension is adopted by setting $\sigma_x=\sigma_y=\lambda$. In this way, each pixel contributes a local Gaussian component, and the final heatmap is obtained by aggregating all Gaussian responses over the mask region. As illustrated in Fig.~\ref{fig:GPM}, different values of $\lambda$ produce Gaussian heatmaps at different spatial scales.

\subsection{Training Objective with Multi-scale Supervision}
The generated multi-scale Gaussian heatmaps are used to supervise intermediate decoder features. Compared with hard binary labels, these heatmaps encode not only target presence but also the relative spatial importance of pixels around salient instances. This provides richer structural guidance and enables the network to progressively learn the spatial extent, central tendency, and structural distribution of foreground objects.
Accordingly, the Gaussian heatmap loss is introduced as an auxiliary term in the overall training objective. By complementing the classification and mask-based losses, it further improves the quality of intermediate feature learning.

Since USIS requires the model to simultaneously perform object-level recognition and pixel-level segmentation, multiple loss terms are jointly optimized during training. The overall loss function is formulated as
\begin{equation}
\mathcal{L}
=\lambda_{\mathrm{cls}}\mathcal{L}_{\mathrm{cls}}
+\lambda_{\mathrm{obj}}\mathcal{L}_{\mathrm{obj}}
+\lambda_{\mathrm{mask}}\mathcal{L}_{\mathrm{mask}}
+\lambda_{\mathrm{dice}}\mathcal{L}_{\mathrm{dice}}
+\lambda_{\mathrm{gh}}\mathcal{L}_{\mathrm{gh}}.
\end{equation}
Here, $\mathcal{L}_{\mathrm{cls}}$ supervises instance category prediction, $\mathcal{L}_{\mathrm{obj}}$ supervises object-level detection, $\mathcal{L}_{\mathrm{mask}}$ constrains the pixel-wise quality of predicted masks, $\mathcal{L}_{\mathrm{dice}}$ improves mask overlap and alleviates foreground--background imbalance, and $\mathcal{L}_{\mathrm{gh}}$ serves as an auxiliary term for guiding intermediate representations with multi-scale Gaussian heatmaps. The weighting coefficients $\lambda_{\mathrm{cls}}$, $\lambda_{\mathrm{obj}}$, $\lambda_{\mathrm{mask}}$, $\lambda_{\mathrm{dice}}$, and $\lambda_{\mathrm{gh}}$ control the relative contributions of the corresponding loss terms.

\section{Implementation Details}
The proposed USIS-PGM model is implemented in PyTorch. All evaluations are conducted on an Ubuntu 20.04 platform with an NVIDIA GeForce RTX 3090 GPU and CUDA support.
We use the USIS16K dataset for training and evaluation. In this dataset, the underwater images and their corresponding depth maps are uniformly resized to $640 \times 480 \times 3$. To ensure effective model training, the USIS-PGM is trained for a total of 500{,}000 iterations with an initial learning rate of $5 \times 10^{-5}$ and a weight decay of 0.0001.
The weights for the classification loss, mask loss, Dice loss, and Gaussian heatmap loss are set to $\lambda_{\mathrm{cls}}=\lambda_{\mathrm{obj}}=\lambda_{\mathrm{mask}}=\lambda_{\mathrm{dice}}=\lambda_{\mathrm{gh}}=0.2$.

\section{Evaluation and Analysis}

\subsection{Evaluation Protocol}
The proposed USIS-PGM model is comprehensively evaluated on the USIS16K dataset against 12 benchmark methods, including Mask R-CNN~\cite{he2017mask}, Mask Scoring R-CNN~\cite{huang2019mask}, YOLACT~\cite{bolya2019yolact}, SOLO~\cite{wang2020solo}, SOLOV2~\cite{wang2020solov2}, PointRend~\cite{kirillov2020pointrend}, DetectoRS~\cite{qiao2021detectors}, SCNet~\cite{vu2021scnet}, SparseInst~\cite{Cheng2022SparseInst}, ConvNeXt~\cite{liu2022convnet}, WaterMask~\cite{lian2023watermask}, and USIS-SAM~\cite{lian2024diving}. For fairness, all competing methods are evaluated under the same dataset protocol and standard evaluation settings. Furthermore, ablation studies are conducted to verify the effectiveness of the proposed PGM-based multi-scale supervision strategy.

\subsection{Qualitative Evaluation}
To provide a more intuitive comparison, Fig.~\ref{fig:qualitative_com} presents qualitative results of the proposed USIS-PGM model and 12 benchmark methods on underwater images from the USIS16K dataset. The visual comparisons show that USIS-PGM generates more complete and accurate salient instance masks, especially in challenging scenarios involving complex backgrounds, ambiguous object boundaries, and multiple underwater objects.

In particular, the masks predicted by USIS-PGM exhibit more precise boundary delineation, fewer missing regions, and clearer separation between adjacent instances. These advantages are especially evident in scenes containing multiple salient objects or thin structures, where benchmark methods often suffer from incomplete masks, false positives, or confusion between neighboring instances. As a result, USIS-PGM produces masks that are not only more accurate but also more spatially coherent. These qualitative observations confirm the effectiveness and superiority of the proposed USIS-PGM.

\subsection{Quantitative Evaluation}
To quantitatively evaluate the performance of the proposed USIS-PGM model, we compare it with 12 benchmark methods on the USIS16K dataset. As shown in Table~\ref{tab:competitors2}, USIS-PGM achieves the best overall mAP of 0.816, outperforming all competing methods, including the underwater-specific model USIS-SAM, which attains an mAP of 0.810. Moreover, USIS-PGM obtains the highest $AP_{75}$ of 0.882 and the best $AP_S$ of 0.568, indicating that the proposed model is particularly effective at producing accurate masks and handling small salient instances under challenging underwater conditions. USIS-PGM also achieves a competitive $AP_M$ of 0.670 and an $AP_L$ of 0.814, demonstrating robust performance across different object scales. Although its $AP_{50}$ is not the highest among all compared methods, its superior results on the more stringent metrics mAP and $AP_{75}$ suggest better overall localization quality and mask accuracy. 

\begin{figure}[h]
    \centering
    \includegraphics[width=0.49\textwidth]{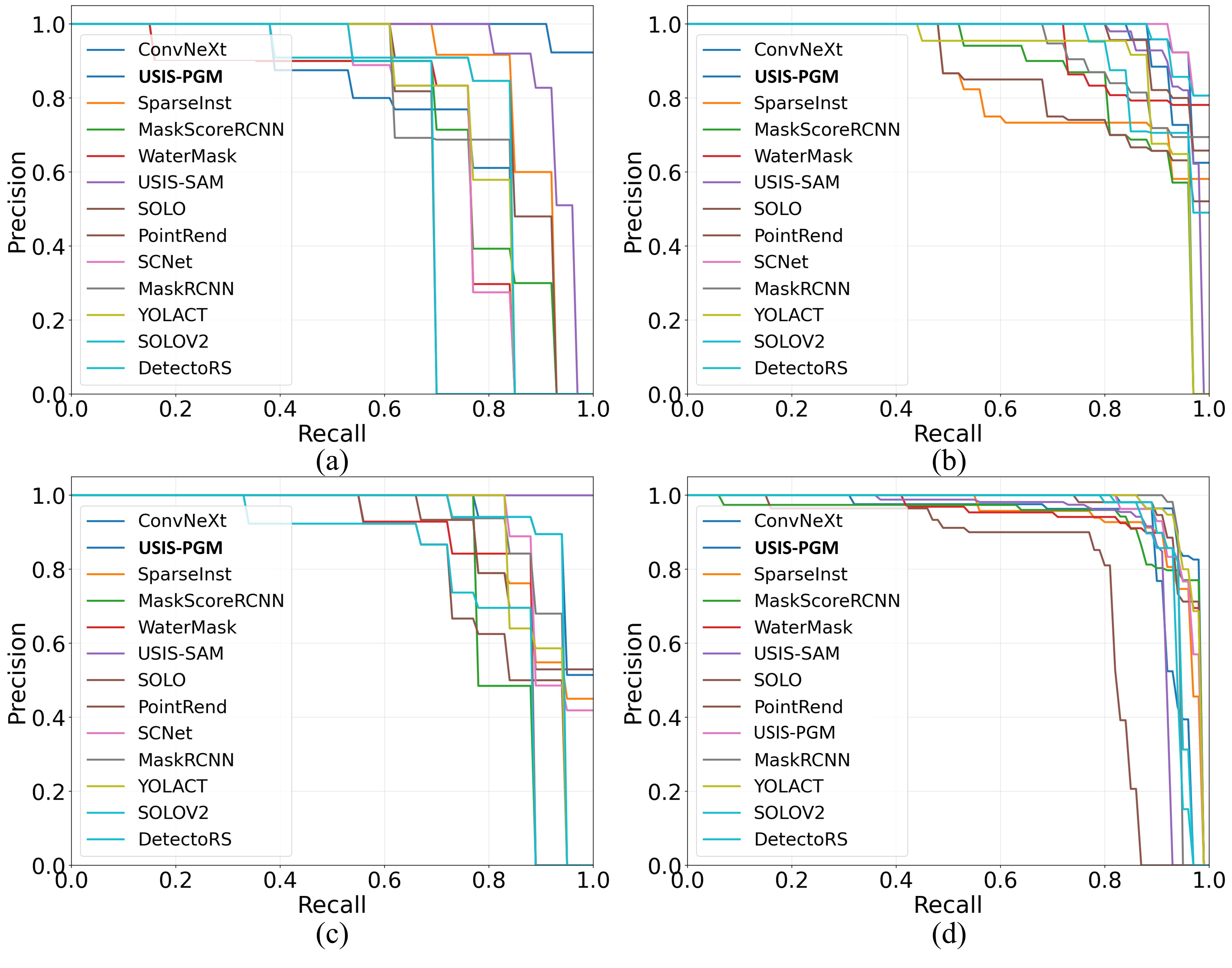}
   \caption{ PR curve performance of the proposed USIS-PGM model and 12 benchmark methods on four representative object categories, i.e., (a) Diver,
(b) Swimmer, (c) Plastic cup, and (d) Sea chest grating.}
\label{fig:PR}
\end{figure}

Fig.~\ref{fig:PR} presents the precision--recall (PR) curves of 12 benchmark methods and the proposed USIS-PGM model on four randomly selected object categories from the USIS16K dataset. It can be observed that USIS-PGM outperforms most competing methods on three categories, namely Diver, Swimmer, and Plastic Cup, demonstrating its strong category-level detection and segmentation capability. Although there is still room for improvement in certain categories, the overall PR-curve performance further confirms the effectiveness of the proposed model. In addition, USIS-SAM and WaterMask, as underwater-specific methods, outperform most benchmark methods originally developed for terrestrial environments, highlighting the importance of domain-specific design for underwater visual perception.
\begin{table}[t]
\begin{center}
\caption{Quantitative evaluation results of the USIS-PGM model.}
\resizebox{\linewidth}{!}{
\begin{tabular}{l|c|c|c|c|c|c}
\toprule
\rowcolor[gray]{0.7}
\textbf{Methods} & $mAP (\uparrow)$ & $AP_{50} (\uparrow)$ & $AP_{75} (\uparrow)$ & $AP_{S} (\uparrow)$ & $AP_{M} (\uparrow)$ & $AP_{L} (\uparrow)$ \\
\hline
Mask R-CNN~\cite{he2017mask} & .736 & .900 & .817 & .400 & .604 & .745 \\
\rowcolor[gray]{0.9}
Mask Scoring R-CNN~\cite{huang2019mask} & .716 & .885 & .798 & .450 & .515 & .726 \\
YOLACT~\cite{bolya2019yolact} & .754 & .904 & .826 & .450 & .545 & .760 \\
\rowcolor[gray]{0.9}
SOLO~\cite{wang2020solo} & .595 & .763 & .659 & .119 & .327 & .606 \\
SOLOV2~\cite{wang2020solov2} & .717 & .859 & .771 & .450 & .482 & .726 \\
\rowcolor[gray]{0.9}
PointRend~\cite{kirillov2020pointrend} & .763 & .894 & .827 & .500 & .659 & .774 \\
DetectoRS~\cite{qiao2021detectors} & .732 & .938 & .872 & .433 & .664 & .788 \\
\rowcolor[gray]{0.9}
SCNet~\cite{vu2021scnet} & .752 & .910 & .822 & .475 & .630 & .759 \\
SparseInst~\cite{Cheng2022SparseInst} & .730 & .850 & .769 & .450 & .449 & .740 \\
\rowcolor[gray]{0.9}
ConvNeXt~\cite{liu2022convnet} & .785 & .953 & .879 & .551 & .688 & .792 \\
Watermask~\cite{lian2023watermask} & .727 & .868 & .793 & .417 & .570 & .736 \\
\rowcolor[gray]{0.9}
USIS-SAM~\cite{lian2024diving} & .810 & .908 & .871 & .450 & .643 & \textbf{.818} \\
\textbf{USIS-PGM (ours)} & \textbf{.816} & .912 & \textbf{.882} & \textbf{.568} & \textbf{.670} & .814 \\
\toprule
\end{tabular}}
\label{tab:competitors2}
\end{center}
\end{table}

Overall, the qualitative and quantitative results consistently demonstrate the effectiveness and superiority of the proposed USIS-PGM model.

\begin{table}[t]
\begin{center}
\caption{Ablation study of the PGM-based multi-scale supervision.}
\resizebox{\linewidth}{!}{
\begin{tabular}{l|c|c|c|c|c|c}
\toprule
\rowcolor[gray]{0.7}
\textbf{Methods} & $mAP (\uparrow)$ & $AP_{50} (\uparrow)$ & $AP_{75} (\uparrow)$ & $AP_{S} (\uparrow)$ & $AP_{M} (\uparrow)$ & $AP_{L} (\uparrow)$ \\
\hline
USIS-PGM (-PGM)   & .798 & .889 & .859 & .540 & .549 & .809 \\
USIS-PGM (ours)   & \textbf{.816} & \textbf{.912} & \textbf{.882} & \textbf{.568} & \textbf{.670} & \textbf{.814} \\
\toprule
\end{tabular}}
\label{tab:ablation}
\end{center}
\end{table}

\subsection{Ablation Study of PGM-based Multi-scale Supervision}
To further validate the effectiveness of the proposed PGM-based multi-scale supervision strategy, an ablation study was conducted. Specifically, the full USIS-PGM model is compared with a variant without PGM-based multi-scale supervision, i.e., USIS-PGM(-PGM), in order to isolate the contribution of the proposed PGM-based multi-scale supervision strategy. As shown in Table~\ref{tab:ablation}, introducing PGM brings consistent improvements across all evaluation metrics. In particular, mAP increases from 0.798 to 0.816, while $AP_{50}$ and $AP_{75}$ improve from 0.889 to 0.912 and from 0.859 to 0.882, respectively. In terms of object scale, $AP_S$, $AP_M$, and $AP_L$ increase from 0.540 to 0.568, from 0.549 to 0.670, and from 0.809 to 0.814, respectively. Notably, the largest gain is observed on medium-sized objects, indicating that the proposed PGM-based multi-scale supervision is particularly effective in improving instance localization and structural modeling at this scale.

\subsection{Applications in Underwater Object Reconstruction}
\begin{figure}[t]
    \centering
    \includegraphics[width=0.45\textwidth]{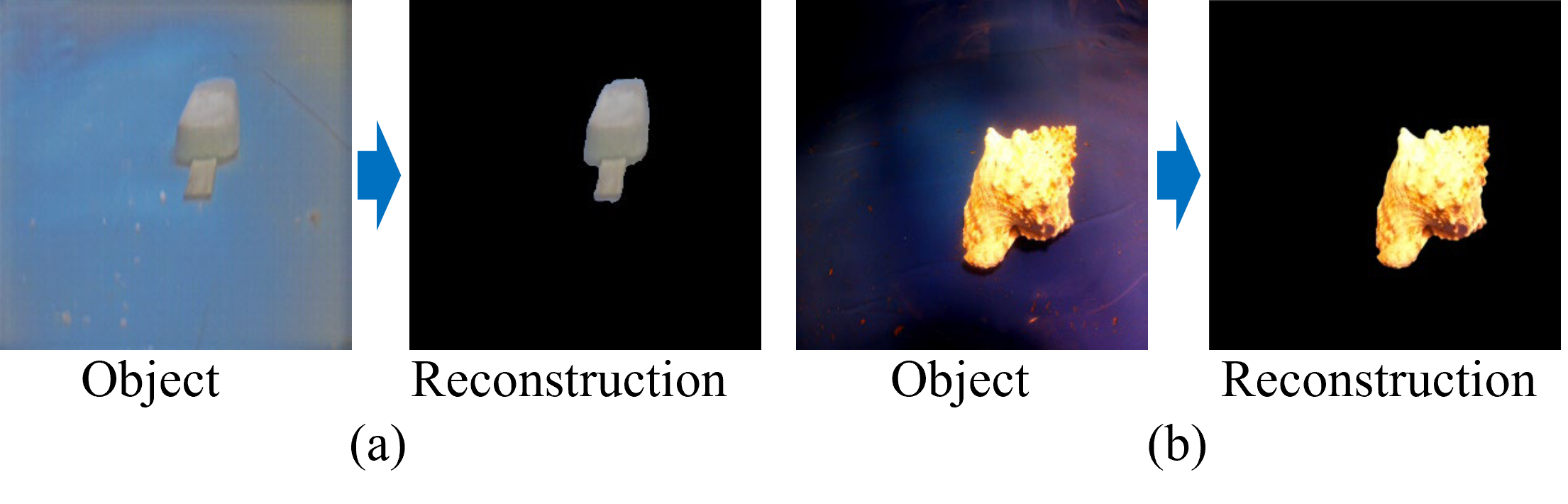}
   \caption{3D reconstruction results of underwater objects, including (a) sacrificial anodes and (b) sea snails, obtained using USIS-PGM and COLMAP~\cite{schonberger2016structure}.}
\label{fig:reconstruction}
\end{figure}

To further demonstrate the practical applicability of USIS-PGM, we integrate the model into an underwater object reconstruction pipeline. Specifically, USIS-PGM is first employed to predict masks for underwater salient instances, such as sacrificial anodes and sea snails, from raw underwater images. The resulting segmentation masks are then used as foreground priors in COLMAP~\cite{schonberger2016structure} for reconstructing the 3D geometry of the underwater objects. As shown in Fig.~\ref{fig:reconstruction}, accurate salient instance segmentation effectively reduces background interference and improves the structural completeness of the reconstructed underwater objects.

\section{Conclusion}
This paper presented a novel USIS-PGM model for USIS. To address the challenges arising from underwater image degradation, the proposed USIS-PGM model integrates frequency-aware boundary enhancement, dynamic feature reweighting, and a Transformer-based instance activation module into a unified single-stage architecture. Furthermore, a multi-scale Gaussian heatmap supervision strategy based on Photometric Gaussian Mixture was introduced to guide intermediate decoder features, thereby promoting more discriminative and spatially coherent learning of salient instance localization and mask structure. Extensive evaluations on benchmark datasets demonstrated that USIS-PGM outperforms existing benchmark methods. Its practical applicability was further demonstrated through underwater object reconstruction experiments. In future work, we will investigate more effective and efficient USIS model for underwater robotic visual perception.
\bibliographystyle{ieeetr} 
\bibliography{ref} 

\end{document}